\documentclass[twocolumn,letterpaper]{IEEEAerospaceCLS}

\usepackage{color,array}
\usepackage{amsmath,amssymb,amsfonts,mathtools,physics,bm,bbm,dsfont,siunitx}
\AtBeginDocument{\RenewCommandCopy\qty\SI}

\usepackage{booktabs}
\usepackage{multirow}
\usepackage{multicol}
\usepackage{tabularx}
\usepackage{dcolumn}

\usepackage{graphicx}
\usepackage{subcaption}
\usepackage{float}
\usepackage{dblfloatfix}

\usepackage{xcolor}
\usepackage{textcomp}

\usepackage{algorithmic}

\usepackage[colorlinks=true, urlcolor=blue, linkcolor=black, citecolor=black]{hyperref}
\usepackage{cleveref}

\usepackage{cite}
\usepackage{xurl}  %
\usepackage{IEEEtrantools}

\newcommand{\best}[1]{\textbf{#1}}
\newcommand{\second}[1]{\underline{#1}}

\newcommand{\tab}[3]{\begin{tabular}[#1]{#2}#3\end{tabular}}

\renewcommand{\arraystretch}{1.3}

\newif\ifdraftmarks \draftmarksfalse  %

\newcommand{\provon}{\ifdraftmarks\color{gray}\fi}
\newcolumntype{P}{>{\provon}c}

\crefname{figure}{Fig.}{Figs.}
\Crefname{figure}{Figure}{Figures}
\crefname{equation}{Eq.}{Eqs.}
\Crefname{equation}{Equation}{Equations}
\crefname{table}{Table}{Tables}
\Crefname{table}{Table}{Tables}
\crefname{section}{Section}{Sections}
\Crefname{section}{Section}{Sections}

\graphicspath{{figures/}}

\newcommand{\hdrstrut}{\rule{0pt}{2.3ex}}

\begin{document}
\bstctlcite{IEEEexample:BSTcontrol}  %

\title{Evaluation of Front-End Feature Matching for Lunar Visual SLAM}

\author{%
    Guillem Casadesus Vila and Grace Gao\\
    Department of Aeronautics and Astronautics\\
    Stanford University\\
    Stanford, CA, United States\\
    \{guillemc,gracegao\}@stanford.edu
}

\maketitle

\thispagestyle{plain}
\pagestyle{plain}

\begin{abstract}
Lunar rovers will primarily rely on cameras for localization until navigation satellite services are deployed, and will continue to need them afterwards for relative navigation, hazard detection, and surface operations. Visual odometry (VO) and Simultaneous Localization and Mapping (SLAM) mostly estimate motion by matching features between frames. However, feature matchers are developed on urban, indoor and driving scenes, planetary studies compare them on orbital registration or within complete SLAM systems, and real lunar imagery with reference poses is scarce, so matching between rover frames on low-texture, harshly lit terrain, and its effect on VO, has not been compared. We benchmark eleven classical and learned detector--matcher combinations on nine synthetic, lunar-testbed, and terrestrial-analogue datasets, and a feature tracker, scoring pairwise pose accuracy against baseline and the trajectory error of the resulting VO. Under a common rotation-only metric, the SuperPoint pairings outperform SIFT with ratio-tested FLANN matching on both lunar testbeds and one analogue site, with smaller or unresolved differences on synthetic data. When rotation and translation direction are scored jointly, this SIFT pairing exceeds both SuperPoint pairings on four synthetic datasets, so synthetic results alone do not establish relative performance on real rover imagery. In stereo VO, the SuperPoint pairings are not consistently more accurate than SIFT, but only they keep tracking on all eight sequential datasets. Pairings with classical detectors depend on lighting while SuperPoint changes little: a low spotlight makes them more accurate than diffuse light on the same testbed views, consistent with more contrast and detected keypoints. SuperPoint with LightGlue is the fastest learned combination (16\,ms per frame), and DINOv2 retrieval proposals still need geometric verification.
\end{abstract}

\tableofcontents

\section{Introduction}
\label{sec:introduction}

Future lunar rovers will drive much farther and faster than any planetary rover so far. Artemis rovers must navigate tens of kilometers from base camps, robotic science rovers at up to 30~cm/s, and the Endurance mission concept requires a rover to traverse over 2700~km of the South Pole--Aitken basin at a nominal 1~km/h~\cite{bhamidipati_trade_2026}, whereas Perseverance has averaged about 2.6~cm/s when driving autonomously~\cite{toupet_enav_2026}. Planetary rovers have so far tracked their motion from the start of each drive and received absolute position updates from operators on Earth, which is expected to limit single autonomous drives to a few hundred meters~\cite{daftry_lunarnav_2023}. Perseverance's recent onboard global localization also relies on cameras~\cite{nash_censible_2024, nasa_perseverance_mgl_2026}. Cameras will thus be the main localization sensor until LunaNet navigation satellite services~\cite{israel_lunanet_2020} are deployed, and will remain essential afterwards for relative navigation, mapping, and hazard detection.

Rovers will rely on visual odometry (VO), which estimates camera motion from consecutive images, and on Simultaneous Localization and Mapping (SLAM), which jointly estimates the trajectory and a map of the environment and bounds accumulated error when it recognizes revisited places~\cite{scaramuzza_visual_2011, cadena_past_2016}. Most VO and SLAM systems, unlike direct methods that use image intensities, detect and match keypoints and estimate the relative pose with random sample consensus (RANSAC) or its variants~\cite{hartley_multiple_2004, barath_magsac_2019, barath_magsacpp_2020}. Correspondence quality therefore affects the trajectory and any map built from it.

Lunar images are difficult to match: regolith is low in texture, and camera optics add distortions and noise. The viewpoint change between matched images also grows when the rover drives faster or skips frames to save compute. At 30~cm/s and 10~Hz a rover advances 3~cm per frame, so VO on every frame matches nearly identical views, while VO on fewer frames matches views up to meters apart. Loop closure instead matches views of the same place taken from a different position and heading, which is expected to be harder on terrain with few distinctive landmarks.

Most matching studies and benchmarks report a single score per dataset~\cite{sarlin_superglue_2020, sun_loftr_2021, jin_imw_2021}, which hides how matching degrades as the viewpoint change grows, and synthetic data alone cannot show whether performance transfers to real imagery. Matching and visual place recognition are also benchmarked separately, although SLAM needs both: retrieval proposes a revisit, and matching verifies it.

We evaluate the correspondences used for VO and for loop closure with the same front ends. Our contributions are:
\begin{itemize}
  \item A benchmark of eleven detector--matcher combinations on nine synthetic, lunar-testbed, and terrestrial-analogue datasets, and a Kanade--Lucas--Tomasi (KLT) tracker, with pose error characterized by the viewpoint change between the images, and the computational cost of the main front ends.
  \item A comparison of classical and learned matchers under camera effects and three lightings, and of whether their relative performance on synthetic imagery carries over to real imagery, scored by rotation alone and by rotation and translation direction jointly.
  \item The operating range of each combination along the baseline, its correlation with the absolute trajectory error (ATE) of VO as a percentage of path length (ATE/L), and a test of whether pairwise matching suffices for loop closure.
\end{itemize}

The code, configurations, pair lists and exclusion masks will be released with a future version of the LuPNT simulator~\cite{casadesusvila_lupnt_2025}.

\section{Related Work}
\label{sec:related}

This section reviews VO and SLAM works, feature detection and matching, their use in space applications, existing matching benchmarks, and place recognition for loop closure.

\subsection{Visual Odometry and SLAM}

Feature-based VO estimates the relative motion from keypoints matched between frames, after rejecting outliers~\cite{scaramuzza_visual_2011, fraundorfer_visual_2012}. This stage is the front end of SLAM~\cite{cadena_past_2016}, and visual SLAM methods are classified as sparse, semi-dense, or dense~\cite{macario_barros_comprehensive_2022}. We evaluate the sparse front end alone.

VO benchmarks score whole trajectories rather than image pairs~\cite{sturm_benchmark_2012, geiger_kitti_2012}. In VO systems, viewpoint change typically enters through the keyframe rule. ORB-SLAM inserts a keyframe when, among other conditions, the current frame tracks fewer than 90\,\% of the reference keyframe's points~\cite{mur-artal_orb-slam_2015}, SVO when the camera has moved more than 12\,\% of the average scene depth~\cite{forster_svo_2014}, and DSO when a weighted sum of optical flow, rotation-compensated flow, and exposure change exceeds a threshold~\cite{engel_direct_2018}. We likewise describe pair difficulty by viewpoint change, through overlap and triangulation angle.

Direct and semi-direct methods such as LSD-SLAM~\cite{engel_lsdslam_2014}, DSO, and SVO minimize photometric error instead of matching descriptors (SVO aligns sparse patches directly), so they have no detector--matcher stage to exchange. LDSO~\cite{gao_ldso_2018} adds corners with ORB descriptors to DSO for loop closure. Our KLT baseline is the closest member of this family that yields correspondences. End-to-end systems such as DROID-SLAM~\cite{teed_droidslam_2021}, DPVO~\cite{teed_dpvo_2023}, and TartanVO~\cite{wang_tartanvo_2021} have no keypoint detector or descriptor matcher to exchange.

\subsection{Feature Detection and Matching}

Classical methods hand-craft detection and description: SIFT~\cite{lowe_distinctive_2004} uses gradient histograms, while ORB~\cite{rublee_orb_2011}, AKAZE~\cite{alcantarilla_akaze_2013}, and BRISK~\cite{leutenegger_brisk_2011} use cheaper binary descriptors. ORB is the front end of ORB-SLAM3~\cite{campos_orb-slam3_2021}. Learned methods replace one or both steps: Key.Net~\cite{barroso-laguna_keynet_2019} learns only the detector, while SuperPoint~\cite{detone_superpoint_2018}, D2-Net~\cite{dusmanu_d2-net_2019}, R2D2~\cite{revaud_r2d2_2019}, DISK~\cite{tyszkiewicz_disk_2020}, ALIKED~\cite{zhao_aliked_2023}, and XFeat~\cite{potje_xfeat_2024} learn both. We evaluate the four classical detectors and SuperPoint, the only learned one with released weights for both SuperGlue and LightGlue.

Descriptors are matched by nearest-neighbor search, exhaustive or approximate as in FLANN~\cite{muja_fast_2009}, and filtered with Lowe's ratio test~\cite{lowe_distinctive_2004} or a mutual cross-check. Learned matchers reason over both keypoint sets jointly: SuperGlue~\cite{sarlin_superglue_2020} uses graph attention and Sinkhorn iterations~\cite{sinkhorn_concern_1967}, and LightGlue~\cite{lindenberger_lightglue_2023} reduces its cost with early exit and point pruning. Both are trained on MegaDepth~\cite{li_megadepth_2018}, LightGlue after pre-training on synthetic homographies. We use them without retraining, so our results include this domain gap.

Detector-free matchers~\cite{xu_matching_survey_2024} such as LoFTR~\cite{sun_loftr_2021}, ASpanFormer~\cite{chen_aspanformer_2022}, MatchFormer~\cite{wang_matchformer_2022}, DKM~\cite{edstedt_dkm_2023}, and RoMa~\cite{edstedt_roma_2024}, and 3D foundation models such as DUSt3R~\cite{wang_dust3r_2024} and MASt3R~\cite{leroy_mast3r_2024}, do not separate into detector and matcher stages, so we leave them out.

\subsection{Feature Matching for Space Applications}

Stereo VO ran for two years on the Mars Exploration Rovers~\cite{maimone_two_2007}. Localization against an orbital map, as in the Mars~2020 Lander Vision System~\cite{johnson_mrl_2022} and crater matching for lunar rovers~\cite{daftry_lunarnav_2023}, gives a global fix that complements VO. In a simulated trade study for the Endurance concept, crater-based localization degraded at low sun and with computed rather than ground-truth stereo depth, while dense intensity registration against orbital imagery stayed within 3~m in every case~\cite{bhamidipati_trade_2026}.

The closest precedents evaluate features inside complete SLAM systems. One study replaces ORB with SuperPoint in ORB-SLAM3~\cite{mollica_slam_benchmark_2023}, and another evaluates visual SLAM in planetary-like and agricultural environments~\cite{romerobautista_evaluation_2024}. Neither varies the matcher, characterizes pairs by viewpoint change, or scores loop closure separately. Matchers have also been compared on Chandrayaan-2 orbital imagery~\cite{makharia_comparative_2025}, a cross-sensor registration task rather than matching between rover frames. DINO-VO~\cite{azhari_dinovo_2025} builds a VO front end on DINOv2 features and has been evaluated only on terrestrial datasets.
Two lunar systems from our group use a SuperPoint and LightGlue front end~\cite{dai_dem_anchoring_2026, dai_fullstack_lac_2026}, and this benchmark tests that choice.

\subsection{Benchmarking Feature Matching}

The Image Matching Challenge~\cite{jin_imw_2021} made recovered pose the primary measure of matching quality. SuperGlue~\cite{sarlin_superglue_2020} and LoFTR~\cite{sun_loftr_2021} report it as Pose AUC, the area under the curve (AUC) of the cumulative pose error, which we adopt. HPatches~\cite{balntas_hpatches_2017} scores descriptors on planar scenes, which have no parallax.

RUBIK~\cite{loiseau_rubik_2025} bins nuScenes pairs by overlap, scale ratio, and the triangulation angle $\bar\alpha$ we also report, but covers one terrestrial driving dataset and scores image pairs only. We apply the same description to lunar and analogue data, sample pairs on a grid of metric translations as a rover would space its frames, and add downstream VO and loop closure.

\subsection{Visual Place Recognition and Loop-Closure Evaluation}

Visual place recognition (VPR) benchmarks score retrieval~\cite{zaffar_vprbench_2021, berton_deepvgbench_2022} but not the pose recovered once a place is found. Revisits are defined by a fixed radius~\cite{arandjelovic_netvlad_2016}, a radius with a viewing-angle limit~\cite{warburg_msls_2020}, or the visual overlap of the two views~\cite{schubert_vprtutorial_2023, leyva-vallina_graded_2023}. We use a radius combined with a minimum arc-length separation.

The closest planetary-analogue loop-closure work is MPRF~\cite{encinargonzalez_mprf_2025}, which combines DINOv2 retrieval with LiDAR-based pose estimation on S3LI sequences including the two we use. To our knowledge, no prior work evaluates the same camera-only front ends for both VO and loop-closure verification across synthetic, lunar-testbed, and terrestrial-analogue data, and relates the pairwise results to VO accuracy.

\section{Problem Formulation and Datasets}
\label{sec:problem_data}

This section states the problem, defines how we measure viewpoint change, and describes the nine datasets and how we sample image pairs.

\begin{table*}[b]
  \centering
  \caption{\bfseries\boldmath Datasets used for evaluation. GNSS: global navigation satellite system.}
  \label{tab:datasets}
  \renewcommand{\arraystretch}{1.1}
  \footnotesize
  \begin{tabularx}{\textwidth}{|l|l>{\raggedright\arraybackslash}Xr|ll|}
    \hline
    \hdrstrut \textbf{\boldmath Dataset} & \textbf{\boldmath Type} & \textbf{\boldmath Conditions} & \textbf{\boldmath Frames} &
    \textbf{\boldmath Reference depth} & \textbf{\boldmath Ground-truth pose} \\
    \hline\hline
    LuPNT (base)   & Synthetic & Sun elevation $15^\circ$ & 4,281 & Rendered & Simulator \\
    LuPNT (cam.)   & Synthetic & Camera effects & 4,276 & Rendered & Simulator \\
    LuPNT (high sun)  & Synthetic & High sun, elevation $45^\circ$ & 4,281 & Rendered & Simulator \\
    LuSNAR (5)     & Synthetic & Scene 5 & 1,599 & Rendered & Simulator \\
    LuSNAR (9)     & Synthetic & Scene 9 & 2,087 & Rendered & Simulator \\
    PolarTraverse  & Lunar testbed & Regolith testbed & 11 stations & Laser scan & COLMAP registered to laser scan \\
    LunaPolaris    & Lunar testbed & Ceiling lights (seq.\,01) & 2,244 & Lidar & Motion capture \\
    S3LI (Etna)    & Terrestrial analogue & Volcanic terrain & 20,833 & Stereo & Differential GNSS, gyroscope \\
    S3LI (Vulcano) & Terrestrial analogue & Volcanic terrain & 12,191 & Stereo & Differential GNSS, gyroscope \\
    \hline
  \end{tabularx}
\end{table*}

\subsection{Problem Formulation}
\label{sec:problem}

VO chains relative poses between consecutive frames, and SLAM verifies each revisit with the same pairwise step: given two frames $I_i, I_j$, detect keypoints, match them, and estimate the relative pose $(\mathbf{R}_{ij}, \mathbf{t}_{ij})$. How well a combination does this depends on how different the two views are, so we score it against viewpoint change. We ask which combination recovers the relative pose most accurately on lunar and analogue imagery and over what baseline, whether its relative performance on synthetic data carries over to real data, and whether pairwise results carry over to VO and revisit verification.

\subsection{Describing Viewpoint Change}
\label{sec:viewpoint}

We describe viewpoint change by the overlap ratio $\omega$ and the triangulation angle $\bar\alpha$, computed from reference pose and depth as in the RUBIK benchmark~\cite{loiseau_rubik_2025}. Let $T^*_k = [\mathbf{R}^*_k \mid \mathbf{t}^*_k]$ be the ground-truth camera-from-world transform of frame $I_k$, $\mathbf{c}^*_k = -\mathbf{R}^{*\top}_k\mathbf{t}^*_k$ its camera center, and $T^*_{ij} = T^*_j (T^*_i)^{-1} = [\mathbf{R}^*_{ij} \mid \mathbf{t}^*_{ij}]$ the relative pose. For a pair, $\mathcal{V}_i$ and $\mathcal{V}_j$ are the non-sky pixels with valid depth. A pixel $\mathbf{u}\in\mathcal{V}_i$ with depth $D_i(\mathbf{u})$ is back-projected with intrinsics $\mathbf{K}$ to $\mathbf{p} = D_i(\mathbf{u})\,\mathbf{K}^{-1}\tilde{\mathbf{u}}$, moved to camera $j$ as $\mathbf{p}' = \mathbf{R}^*_{ij}\mathbf{p} + \mathbf{t}^*_{ij}$, and projected to a pixel $\mathbf{u}'$. It is \emph{co-visible} if $\mathbf{p}'$ lies in front of camera $j$, $\mathbf{u}'$ lands on a valid pixel of $I_j$, and the depth of $\mathbf{p}'$ matches $D_j(\mathbf{u}')$ within 5\,\%, which rejects occlusions. With $\mathcal{C}_{i\to j}\subseteq\mathcal{V}_i$ and $\mathcal{C}_{j\to i}\subseteq\mathcal{V}_j$ the co-visible pixels in each direction, the overlap ratio is
\begin{equation}
  \omega = \frac{|\mathcal{C}_{i\to j}| + |\mathcal{C}_{j\to i}|}{|\mathcal{V}_i| + |\mathcal{V}_j|}.
  \label{eq:omega}
\end{equation}
Overlap selects training pairs in D2-Net~\cite{dusmanu_d2-net_2019} and test pairs in SuperGlue~\cite{sarlin_superglue_2020}, and we use it as the primary axis.

Overlap does not measure parallax, so we also compute the median triangulation angle $\bar\alpha$~\cite{hartley_multiple_2004}. For a co-visible point $\mathbf{X}$,
\begin{equation}
  \alpha(\mathbf{X}) = \arccos\!\left(\frac{(\mathbf{X}-\mathbf{c}^*_i)\cdot(\mathbf{X}-\mathbf{c}^*_j)}{\|\mathbf{X}-\mathbf{c}^*_i\|\,\|\mathbf{X}-\mathbf{c}^*_j\|}\right),
  \label{eq:triangulation_angle}
\end{equation}
and $\bar\alpha$ is its median over the co-visible points in both directions. A pure rotation gives $\bar\alpha = 0$. For the axis comparison we also use the scale ratio $\delta$, the median of $\max(d_i,d_j)/\min(d_i,d_j)$, where $d_i$ and $d_j$ are the distances from $\mathbf{X}$ to the two camera centers.

The reference depth differs by tier (Table~\ref{tab:datasets}). On S3LI it comes from RAFT-Stereo~\cite{lipson_raftstereo_2021} and is cut at $Z_\mathrm{max} = 0.1\,f_x B$, where a 1-pixel disparity error causes a 10\,\% depth error ($f_x$ is the focal length in pixels and $B$ the baseline). The cutoff drops distant terrain, which lowers $\omega$ and raises $\bar\alpha$, and on LuSNAR (5) cutting rendered depth at $Z_\mathrm{max}$ closely reproduces the stereo values. Absolute $\omega$ and $\bar\alpha$ are therefore not comparable across tiers, so we compare overlap-binned accuracy only within a tier.

The camera translation $\Delta t$ and rotation $\Delta\theta$ need only the ground-truth poses,
\begin{align}
  \Delta t &= \|\mathbf{c}^*_j - \mathbf{c}^*_i\|_2, \label{eq:delta_t}\\
  \Delta\theta &= \arccos\!\Bigl(\tfrac{\mathrm{tr}(\mathbf{R}^{*\top}_i \mathbf{R}^*_j) - 1}{2}\Bigr).
  \label{eq:delta_theta}
\end{align}
We use them to sample frames and to report per-step accuracy against $\Delta t$, but not as difficulty axes, because the difficulty of a translation depends on scene depth and a rotation changes overlap without adding parallax.

\subsection{Datasets and Ground Truth}
\label{sec:datasets}

\begin{figure*}[b]
  \centering
  \includegraphics[width=\linewidth]{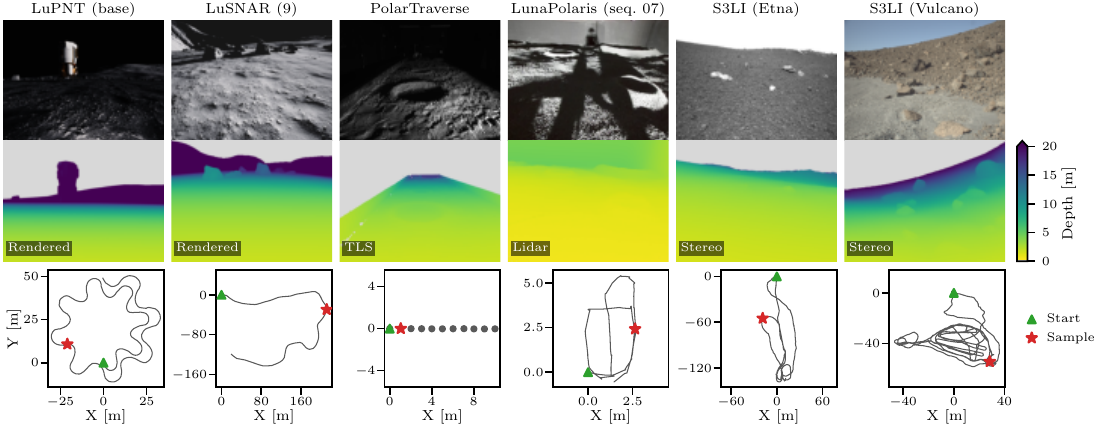}
  \caption{\bfseries\boldmath Sample image, reference depth (sky and invalid pixels gray) and top-down trajectory with the sample frame marked, for six of the nine datasets. LunaPolaris shows sequence 07. TLS: terrestrial laser scan.}
  \label{fig:dataset_overview}
\end{figure*}

We use nine datasets in three tiers: \emph{synthetic} data from LuPNT and LuSNAR~\cite{liu_lusnarlunar_2024}, \emph{lunar testbed} data from the indoor analogue facilities of PolarTraverse~\cite{hansen_polar_2024} and LunaPolaris~\cite{van_der_meer_lunapolaris_2025}, and \emph{terrestrial analogue} data from S3LI, recorded on volcanic terrain~\cite{giubilato_challenges_2022,giubilato_s3livulcano_2026}. Table~\ref{tab:datasets} summarizes them, and Fig.~\ref{fig:dataset_overview} shows examples.

\subsubsection{LuPNT Simulator (Synthetic)}

We use LuPNT~\cite{casadesusvila_lupnt_2025}, an open-source simulator for lunar communications, positioning, navigation, and timing, with its Unreal Engine 5.7 surface rendering. A rover drives over south-polar terrain built from high-resolution elevation models with procedurally generated rocks and craters. We render a stereo camera along one 378\,m path that loops around a lander, in three conditions: \emph{base} (sun elevation $15^\circ$, our reference condition, where the surrounding relief shadows much of the terrain), \emph{camera effects}, and \emph{high sun} (sun elevation $45^\circ$).
The camera-effects run adds depth-of-field blur, chromatic aberration, bloom, and lens flare.

\subsubsection{LuSNAR (Synthetic)}

LuSNAR~\cite{liu_lusnarlunar_2024} provides stereo imagery of nine lunar scenes built in Unreal Engine~4. We use scenes 9 and 5, whose terrain and rendering differ from LuPNT, as two datasets, downsampled to 512 by 512 pixels, close to the real image sizes.

\subsubsection{PolarTraverse (Lunar Testbed)}

This dataset~\cite{hansen_polar_2024} was recorded over a regolith-simulant test bed shaped after south-polar terrain and lit from about $2$--$3^\circ$ elevation, with a stereo bar stopping at 11 stations 1\,m apart. Poses are refined with COLMAP~\cite{schonberger_colmap_2016} and registered to a laser scan of the test bed, which also provides the reference depth. Because these poses come from the images themselves, the PolarTraverse pose scores depend on their accuracy. We use one image set (one camera view, traverse and exposure time) and score its 38 station pairs.

\subsubsection{LunaPolaris (Lunar Testbed)}

LunaPolaris~\cite{van_der_meer_lunapolaris_2025} was recorded by a small rover in the LunaLab facility of the University of Luxembourg, in twelve sequences that repeat one trajectory (27.5\,m evaluated) under three boulder layouts and four lighting setups. We use sequence 01 (ceiling lights), and for the lighting comparison also 07 (a low spotlight) and 10 (the spotlight and a rover headlight). Images come from a ZED2 stereo camera, and the reference depth is a dense lidar map of the arena rendered into the camera.
A hand-eye calibration per sequence locates the onboard lidar, and through the rig calibration the camera, relative to the motion-capture marker. Left and right images are paired by capture time, and the motion-capture pose is interpolated at that time rather than at the time of receipt. We skip frames with motion-capture dropouts. In the small test bed some pairs move under 10\,cm, leaving their reference translation direction poorly defined.

\subsubsection{S3LI (Terrestrial Analogue)}

S3LI contributes two sequences, Etna~\cite{giubilato_challenges_2022} (loops) and Vulcano~\cite{giubilato_s3livulcano_2026} (moon lake), neither with dense ground-truth depth.
Positions come from differential Global Navigation Satellite System (GNSS) receivers. Fixed-ambiguity solutions, accurate to 1 to 10\,cm~\cite{giubilato_challenges_2022, giubilato_s3livulcano_2026}, cover nearly all our Etna pair frames but almost none of Vulcano. We drop frames at GNSS position jumps where positions are used. Orientation is integrated from the gyroscope and aligned to the GNSS east-north-up frame using gravity and GNSS velocity. Its median error per pair, estimated against an independent visual rotation track, is at most $0.31^\circ$, and we check the S3LI results under rotation perturbations of up to three times this error. As heading is uncertain by several degrees, epipolar precision (EP), $\omega$ and $\bar\alpha$ on S3LI depend on the translation direction and are approximate.

\subsection{Pair Sampling}
\label{sec:pair_sampling}

Unless noted, pairwise results use a \emph{mixed pool} of up to 1000 independent pairs per dataset, drawn at frame steps chosen per dataset so that their median translations run from about 0.1 to 8\,m. Because the rover speed varies, one frame step covers a range of translations, so we label each step by the median translation $\Delta t$ of its pairs. On LunaPolaris sequence 07, evaluated at 3.0 rather than 5.0\,Hz, a frame step spans a longer baseline.

\begin{figure}[t]
  \centering
  \includegraphics[width=\linewidth]{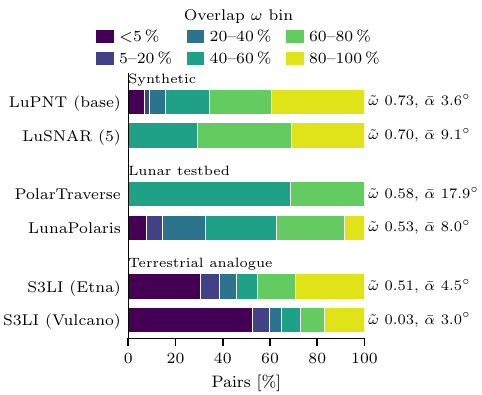}
  \caption{\bfseries\boldmath Fraction of mixed-pool pairs per overlap bin (RUBIK~\cite{loiseau_rubik_2025} bin edges plus a 0--5\,\% bin) on six datasets, with median overlap $\tilde\omega$ and triangulation angle $\bar\alpha$.}
  \label{fig:omega_coverage}
\end{figure}

LuPNT (base) fills every overlap bin, LuSNAR (5) and PolarTraverse lack low-overlap pairs, and LunaPolaris and the S3LI sites have many pairs with almost no overlap (Fig.~\ref{fig:omega_coverage}). Along a traverse, translation, rotation and parallax grow jointly, so these axes are strongly coupled in synthetic data.

\section{Methods}
\label{sec:methods}

We evaluate the front end alone and use no pose-graph or bundle-adjustment back end, so every trajectory is open-loop.

\subsection{Detectors and Matchers}
\label{sec:methods_models}

\label{sec:detectors}

The detectors are SIFT~\cite{lowe_distinctive_2004}, ORB~\cite{rublee_orb_2011}, AKAZE~\cite{alcantarilla_akaze_2013}, BRISK~\cite{leutenegger_brisk_2011}, and SuperPoint~\cite{detone_superpoint_2018}, a convolutional network trained by homographic adaptation. SIFT and SuperPoint have real-valued descriptors, the others binary ones.

\label{sec:matchers}

FLANN~\cite{muja_fast_2009, muja_scalable_2014} searches approximate nearest neighbors with randomized $k$-d trees for real-valued and OpenCV's locality-sensitive hashing index for binary descriptors, and BruteForce searches exhaustively. As is standard, FLANN matches pass Lowe's ratio test at $0.75$ and BruteForce matches a mutual cross-check. Because FLANN is randomized, its results report the median over five seeds. SuperGlue~\cite{sarlin_superglue_2020} and LightGlue~\cite{lindenberger_lightglue_2023} keep only confident matches, and we use their MegaDepth~\cite{li_megadepth_2018} weights and feed LightGlue RootSIFT descriptors for SIFT.

As a tracking baseline, we pair Shi--Tomasi corners~\cite{shi_good_1994} with Lucas--Kanade tracking~\cite{lucas_iterative_1981} over an image pyramid and a forward-backward check. Features are tracked either directly from $I_i$ to $I_j$, as the matchers see them, or through every intermediate frame (none on PolarTraverse).

SuperPoint (SP) is paired only with the learned matchers, which gives 11 combinations: SIFT, ORB, AKAZE and BRISK each with FLANN and BruteForce, plus SIFT$+$LightGlue, SP$+$SuperGlue and SP$+$LightGlue. To separate detector from matcher, four ablation combinations pair SuperPoint and RootSIFT with FLANN and BruteForce under the same filtering.

\subsection{Visual Odometry Front-End}
\label{sec:vo_slam}
\label{sec:vo_pipeline}

Our frame-to-frame VO back-projects the keypoints of $I_{k-1}$ matched to $I_k$ with their depth, estimates the pose of $I_k$ with OpenCV's~\cite{bradski_opencv_2000} RANSAC~\cite{fischler_ransac_1981} Perspective-n-Point (PnP) solver, and chains it into the trajectory. If too few inliers have metric depth, the pose is rejected and the next frame is matched against the last accepted one. Since a rover has no ground-truth depth, stereo VO, run on every sequential dataset, takes depth from the disparity of left-right matches found with the same matcher. Stereo matches must be one-to-one with a row offset within 2\,px of the frame's median offset, and matches with zero disparity act as rotation constraints. On LuPNT and LuSNAR we also run VO with rendered ground-truth depth, which isolates matching and pose estimation from depth error. The real datasets also differ from the synthetic ones in terrain, so their VO results are not directly comparable. PolarTraverse has no image sequence and is evaluated pairwise only.

\subsection{Evaluation Metrics and Protocol}
\label{sec:metrics}

\subsubsection{Pairwise metrics}
\label{sec:pairwise_metrics}

EP, reprojection precision and Pose AUC follow Glue Factory~\cite{glue_factory_repo} with known intrinsics. A star ($*$) marks ground truth and a hat ($\hat{\cdot}$) estimates.

EP is the fraction of candidate matches that lie within $\tau_\mathrm{epi} = 10^{-3}$ in normalized camera coordinates (about $0.5$\,px on LuPNT) of their epipolar line under the ground-truth essential matrix $\mathbf{E}^* = [\mathbf{t}^*_{ij}]_\times\mathbf{R}^*_{ij}$. Because it scores all candidate matches, a matcher is penalized for wrong matches even when RANSAC would discard them. Reprojection precision is the fraction of candidate matches within a pixel threshold of the correspondence given by reference depth.

For Pose AUC, we estimate the essential matrix from the candidate matches with RANSAC and the five-point solver~\cite{nister_efficient_2004} and decompose it into a rotation $\hat{\mathbf{R}}$ and a translation direction $\hat{\mathbf{t}}$. The rotation error $e_R$ is the angle between $\hat{\mathbf{R}}$ and $\mathbf{R}^*_{ij}$, and the translation error $e_t$ the angle between $\hat{\mathbf{t}}$ and the reference direction, up to sign. As is standard~\cite{sarlin_superglue_2020, sun_loftr_2021, lindenberger_lightglue_2023}, the joint pose error is $e = \max(e_R, e_t)$ on the synthetic tier and PolarTraverse. On S3LI and LunaPolaris the reference direction is not accurate enough, so we score rotation alone, $e = e_R$, and mark these results rotation-only. $\mathrm{AUC}@5^\circ$ is the area under the curve of the fraction of pairs with error below a threshold, for thresholds from $0$ to $5^\circ$, as a percentage, and a pair whose pose cannot be estimated counts as a miss. On LunaPolaris, the 26 pairs that include a motion-capture glitch frame are not scored. For a common metric across datasets, Table~\ref{tab:rotonly} scores rotation alone on all nine.

The \emph{operating range} of a combination on a dataset is the set of frame steps $\Delta t$ at which Pose AUC@$5^\circ$ (as scored on that dataset) is at least $50\,\%$,
\begin{equation}
  \mathcal{R} = \bigl\{\Delta t : \mathrm{AUC}@5^\circ(\Delta t) \geq 50\,\%\bigr\},
  \label{eq:operating_range}
\end{equation}
and its upper end $\Delta t^* = \sup\mathcal{R}$, interpolated log-linearly between steps, is the largest baseline the combination handles. A combination that never reaches $50\,\%$ has no range. PolarTraverse has too few pairs per step for a range.

To tell whether two combinations differ, we compute a bootstrap 95\,\% confidence interval of their difference in Pose AUC@$5^\circ$, scoring both on the same resampled pairs. Because nearby pairs see the same terrain, we resample blocks of neighboring pairs. A difference is unresolved when this interval contains both positive and negative values, so the data cannot tell which is better. For the lighting comparison, we pair frames across sequences by pose, within 0.05\,m and $3^\circ$ (loosely 0.10\,m and $5^\circ$), and Holm-adjust the $p$-values over all tested changes. A qualitative check uses a smaller, screened set of pose-matched pairs.

To choose a reference difficulty axis, we bin the mixed-pool pairs of each synthetic dataset into deciles of $\omega$, $\bar\alpha$, $\delta$, $\Delta t$ and $\Delta\theta$ and compute $\eta^2$, the fraction of per-pair EP variance explained by the decile. We average $\eta^2$ over SIFT+FLANN, AKAZE+FLANN and the SuperPoint pairings.

\subsubsection{Trajectory metrics}
\label{sec:traj_metrics}

We align the estimated VO poses, taken at the rover body origin, to the ground truth with an Umeyama SE(3) fit (no scale). The Absolute Trajectory Error (ATE) is the root-mean-square (RMS) error of the aligned positions, and ATE/L expresses it as a percentage of the ground-truth path length $L$ of the evaluation window (on S3LI, the first 10\,\% of each sequence). As a whole-trajectory error after alignment, ATE/L is not directly comparable to the start-anchored end-point error used in rover operations. A run that diverges to non-finite poses is scored up to that point, with the earlier frames counted as tracked. Path coverage is the fraction of the path along which VO keeps tracking, and a gap of more than 10 frames counts as lost. We report ATE and ATE/L only for runs with at least 90\,\% coverage and mark the others as stopped.

A single VO run changes when its input images change slightly or when it is repeated, so we treat ATE differences below this run-to-run variation as unresolved, a practical rule, not a test of statistical equivalence. Slightly perturbed images move the ATE of SP+LightGlue on LuPNT (base) by up to about $0.26$\,m, our default tolerance. GPU runs and randomized FLANN matching vary more, by up to $0.64$\,m and $1.46$\,m, so their comparisons use the larger spread.

\subsubsection{Revisit and retrieval evaluation}
\label{sec:slam_lc}

A ground-truth revisit is a pair of frames within a radius $R$ of each other and at least $S=20$\,m apart along the path. Consecutive revisit frames within 10\,m along the path form one pass. Passes linked by revisits form a revisit component, which is one revisited place. A dataset's revisits are low-angle if at least half the pairs rotate less than $30^\circ$, and wide-angle otherwise.

Since loop closure relies on the front end to verify a candidate revisit, we score pose recovery on true revisits, not loop detection, using the ground-truth revisit pairs of LuPNT (base) within 2\,m and of LuSNAR (5) within 1\,m. The scores are Pose AUC, the number of candidate matches and the RANSAC inlier fraction.

We evaluate whole-image retrieval on datasets with at least three revisit components (Table~\ref{tab:loop_closure}), using the DINOv2~\cite{oquab_dinov2_2023} class (CLS) token with cosine similarity on evenly spaced frames and radii $R \in \{1,2,5\}$\,m, smaller than the 25\,m of urban benchmarks~\cite{arandjelovic_netvlad_2016, warburg_msls_2020} to match a rover traverse. Each frame queries frames more than 20\,m away along the path, either only earlier ones (causal, as in online loop closure) or both sides (offline). Its top match is correct if the two cameras lie within $R$, and any match for a query without a revisit is a false positive. We report recall@1, the area under the precision-recall curve (AUPRC)~\cite{zaffar_vprbench_2021}, and R@100P, the recall before the first false positive~\cite{schubert_vprtutorial_2023}, once the key loop-closure operating point~\cite{lowry_vprsurvey_2016}.

\subsection{Comparing the Operating Range with VO ATE/L}
\label{sec:prediction}

For each combination and synthetic dataset, we compare $\Delta t^*$ (Eq.~\eqref{eq:operating_range}) with the ATE/L of VO with ground-truth and with stereo depth by a pooled Spearman correlation, excluding combinations without a range. The correlation is in-sample, and its points are not independent because combinations share detectors and matchers and the LuPNT conditions share one path, so we also bootstrap over datasets or combinations.

\subsection{Implementation}
\label{sec:implementation}

Experiments ran on two workstations, one with an AMD Ryzen 9 9950X CPU and an NVIDIA RTX 5090 GPU. Learned methods run on the GPU and classical ones on the CPU. All use published default settings with no per-dataset tuning, so ORB keeps at most 500 keypoints and the other detectors are uncapped. We also test relaxed ORB and AKAZE settings.

We time the FLANN and learned combinations on consecutive LuPNT (base) frames on the RTX 5090 workstation, counting one extraction and one match per frame, without stereo matching or pose estimation.

\section{Results and Discussion}
\label{sec:results}

\begin{table*}[p]
  \centering
  \caption{\bfseries\boldmath Pose AUC@$5^\circ$ [\%] of eleven combinations and, in italics, four ablation pairings. The pose error is the larger of the rotation and translation-direction errors, except on LunaPolaris and S3LI (rotation alone). Bold/underline = best/second among the eleven. FLANN cells: median of five seeds, spread $\leq 4.9$. Unmarked KLT rows track between the two frames (direct) or through every intermediate frame (chained, ``---'' without a sequence). AUC@$5^\circ$: area under the cumulative pose-error curve up to $5^\circ$, FLANN: fast library for approximate nearest neighbors, KLT: Kanade--Lucas--Tomasi, SP: SuperPoint.}
  \label{tab:accuracy}
  \renewcommand{\arraystretch}{1.1}
  \footnotesize
  \begin{tabular}{|ll|ccc|cc|cc|cc|}
\hline
\hdrstrut & & \multicolumn{5}{c|}{\textbf{\boldmath Synthetic}} & \multicolumn{2}{c|}{\textbf{\boldmath Lunar testbed}} & \multicolumn{2}{c|}{\textbf{\boldmath Terrestrial analogue}} \\
\cline{3-7}\cline{8-9}\cline{10-11}
\hdrstrut & & \multicolumn{3}{c|}{\textbf{\boldmath LuPNT}} & \multicolumn{2}{c|}{\textbf{\boldmath LuSNAR}} & \textbf{\boldmath PolarTraverse} & \textbf{\boldmath LunaPolaris} & \multicolumn{2}{c|}{\textbf{\boldmath S3LI}} \\
\cline{3-5}\cline{6-7}\cline{10-11}
\textbf{\boldmath Detector} & \textbf{\boldmath Matcher} & \textbf{\boldmath (base)} & \textbf{\boldmath (cam.)} & \textbf{\boldmath (high sun)} & \textbf{\boldmath (5)} & \textbf{\boldmath (9)} & & & \textbf{\boldmath (Etna)} & \textbf{\boldmath (Vulcano)} \\
\hline\hline
SIFT & FLANN & 58.4 & \second{68.5} & \second{74.9} & \best{67.1} & \best{69.6} & 37.8 & 31.6 & 65.0 & 76.5 \\
SIFT & BruteForce & 43.2 & 53.1 & 55.9 & 51.0 & 53.5 & 28.8 & 15.0 & 47.0 & 69.3 \\
SIFT & LightGlue & 60.4 & \best{70.5} & \best{75.2} & \second{63.2} & \second{66.1} & 44.9 & 29.2 & 69.9 & \best{78.7} \\
ORB & FLANN & 14.4 & 25.7 & 26.9 & 32.8 & 33.6 & 3.7 & 9.9 & 34.4 & 49.8 \\
ORB & BruteForce & 13.9 & 27.7 & 26.9 & 35.9 & 37.4 & 3.4 & 6.7 & 34.4 & 50.1 \\
AKAZE & FLANN & 42.2 & 51.8 & 58.1 & 55.0 & 58.1 & 17.4 & 12.5 & 52.0 & 57.6 \\
AKAZE & BruteForce & 37.5 & 48.3 & 53.4 & 51.7 & 55.0 & 15.4 & 11.0 & 46.9 & 54.9 \\
BRISK & FLANN & 49.8 & 49.5 & 60.8 & 54.5 & 57.0 & 16.6 & 30.4 & 50.8 & 68.6 \\
BRISK & BruteForce & 36.6 & 41.4 & 46.9 & 38.7 & 42.7 & 5.5 & 9.0 & 40.0 & 59.8 \\
SP & SuperGlue & \second{62.1} & 61.1 & 63.4 & 56.1 & 59.0 & \best{93.3} & \best{56.9} & \best{77.0} & \second{77.1} \\
SP & LightGlue & \best{62.2} & 61.0 & 64.0 & 56.4 & 57.3 & \second{93.3} & \second{55.2} & \second{74.8} & 75.5 \\
\hline
\textit{RootSIFT} & \textit{FLANN} & \textit{59.6} & \textit{69.4} & \textit{74.9} & \textit{67.8} & \textit{69.5} & \textit{39.5} & \textit{32.6} & \textit{65.5} & \textit{76.1} \\
\textit{RootSIFT} & \textit{BruteForce} & \textit{43.1} & \textit{53.2} & \textit{54.8} & \textit{51.3} & \textit{53.7} & \textit{30.9} & \textit{17.2} & \textit{47.1} & \textit{70.5} \\
\textit{SP} & \textit{FLANN} & \textit{54.1} & \textit{51.5} & \textit{56.6} & \textit{48.2} & \textit{50.0} & \textit{83.8} & \textit{46.4} & \textit{70.7} & \textit{71.5} \\
\textit{SP} & \textit{BruteForce} & \textit{51.2} & \textit{52.1} & \textit{54.8} & \textit{51.2} & \textit{53.5} & \textit{37.5} & \textit{50.9} & \textit{72.1} & \textit{68.4} \\
\hline
KLT & direct & 34.5 & 39.8 & 41.3 & 53.2 & 55.7 & 6.3 & 29.4 & 35.5 & 55.7 \\
KLT & chained & 75.4 & 76.6 & 78.0 & 84.4 & 84.0 & --- & 61.4 & 69.4 & 74.9 \\
\hline
\end{tabular}

\end{table*}

\begin{table*}[p]
  \centering
  \caption{\bfseries\boldmath Rotation-only Pose AUC@$5^\circ$ [\%] on all nine datasets. Bold/underline = best/second. SIFT+FLANN: median of five seeds. KLT rows, tracked as in Table~\ref{tab:accuracy}, are not marked (``---'': no sequence to chain through). AUC@$5^\circ$: area under the cumulative rotation-error curve up to $5^\circ$, FLANN: fast library for approximate nearest neighbors, KLT: Kanade--Lucas--Tomasi, SP: SuperPoint.}
  \label{tab:rotonly}
  \renewcommand{\arraystretch}{1.1}
  \footnotesize
  \begin{tabular}{|ll|ccc|cc|cc|cc|}
\hline
\hdrstrut & & \multicolumn{5}{c|}{\textbf{\boldmath Synthetic}} & \multicolumn{2}{c|}{\textbf{\boldmath Lunar testbed}} & \multicolumn{2}{c|}{\textbf{\boldmath Terrestrial analogue}} \\
\cline{3-7}\cline{8-9}\cline{10-11}
\hdrstrut & & \multicolumn{3}{c|}{\textbf{\boldmath LuPNT}} & \multicolumn{2}{c|}{\textbf{\boldmath LuSNAR}} & \textbf{\boldmath PolarTraverse} & \textbf{\boldmath LunaPolaris} & \multicolumn{2}{c|}{\textbf{\boldmath S3LI}} \\
\cline{3-5}\cline{6-7}\cline{10-11}
\textbf{\boldmath Detector} & \textbf{\boldmath Matcher} & \textbf{\boldmath (base)} & \textbf{\boldmath (cam.)} & \textbf{\boldmath (high sun)} & \textbf{\boldmath (5)} & \textbf{\boldmath (9)} & & & \textbf{\boldmath (Etna)} & \textbf{\boldmath (Vulcano)} \\
\hline\hline
SIFT & FLANN & 79.5 & 85.2 & \second{89.7} & \best{92.2} & \second{90.2} & 47.3 & 31.6 & 65.0 & 76.5 \\
SIFT & LightGlue & 81.9 & 86.7 & \best{90.3} & 90.9 & 88.4 & 53.2 & 29.2 & 69.9 & \best{78.7} \\
SP & SuperGlue & \best{89.0} & \second{87.2} & 89.3 & 90.8 & \best{90.3} & \second{97.2} & \best{56.9} & \best{77.0} & \second{77.1} \\
SP & LightGlue & \second{88.8} & \best{87.6} & 89.7 & \second{91.1} & 88.4 & \best{97.5} & \second{55.2} & \second{74.8} & 75.5 \\
\hline
KLT & direct & 51.2 & 55.6 & 57.6 & 82.5 & 82.1 & 24.8 & 29.4 & 35.5 & 55.7 \\
KLT & chained & 88.5 & 88.2 & 90.1 & 95.7 & 95.6 & --- & 61.4 & 69.4 & 74.9 \\
\hline
\end{tabular}

\end{table*}

\begin{figure*}[p]
  \centering
  \includegraphics[width=0.90\linewidth]{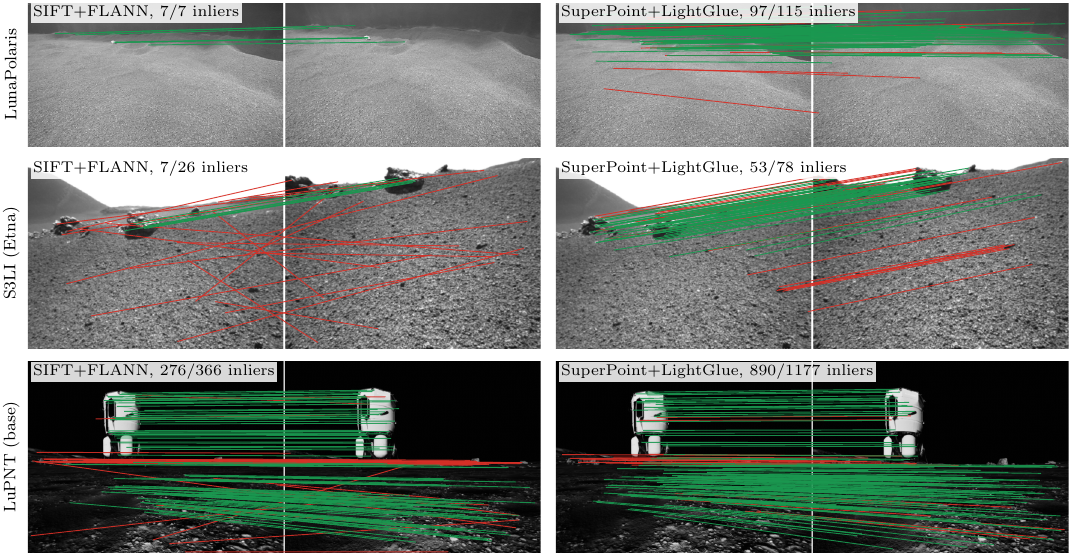}
  \caption{\bfseries\boldmath Matches of SIFT+FLANN (left) and SuperPoint+LightGlue (right) on one pair per dataset, from top: LunaPolaris ($0.34$\,m, $4.0^\circ$), S3LI (Etna) ($2.78$\,m, $29.2^\circ$) and LuPNT (base) ($1.05$\,m, $0.9^\circ$). Green = RANSAC inliers, red = outliers. FLANN: fast library for approximate nearest neighbors, RANSAC: random sample consensus.}
  \label{fig:qual_matches}
\end{figure*}

We first compare pose accuracy across datasets and along the baseline, then camera effects, lighting and cross-dataset transfer, VO and loop closure, and finally the cost of each front end.

In all tables, \best{bold} and \second{underline} mark the best and second-best value, which does not by itself mean the difference is resolved. Unless noted, pairwise metrics average over the mixed pool and VO runs use consecutive evaluated frames. A difference is ``not resolved'' when the data cannot tell which combination is better (Section~\ref{sec:methods}).

\subsection{Accuracy Across Datasets}
\label{sec:results_benchmark}

Fig.~\ref{fig:qual_matches} compares the matches of a classical and a learned combination on one pair from each tier. Both recover the pose on the synthetic pair, but on the two real pairs only the learned one still matches the terrain densely, while the classical one keeps scattered correspondences and fails.

\begin{figure*}[t]
  \centering
  \includegraphics[width=\linewidth]{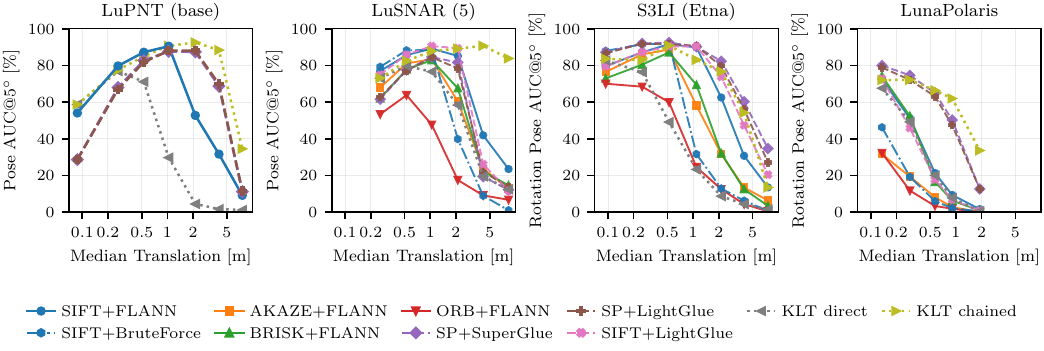}
  \caption{\bfseries\boldmath Pose AUC@$5^\circ$ [\%] against the median translation of each frame step [m, log axis], on LuPNT (base), LuSNAR (5), S3LI (Etna) and LunaPolaris. The two real datasets are scored rotation-only. On the synthetic panels the error includes the translation direction, poorly constrained by the small parallax of the shortest steps, so the AUC first rises with the step. At the widest steps, low overlap also fails the rotation. AUC: area under the curve, FLANN: fast library for approximate nearest neighbors, KLT: Kanade--Lucas--Tomasi, SP: SuperPoint.}
  \label{fig:step_auc}
\end{figure*}

Under a common rotation-only metric (Table~\ref{tab:rotonly}), both SuperPoint pairings outperform SIFT+FLANN on PolarTraverse, LunaPolaris and S3LI (Etna), and by smaller margins on LuPNT (base) and LuPNT (cam.). Elsewhere the difference is not resolved, and SIFT+FLANN is not resolved ahead on any dataset. Under the joint metric of Table~\ref{tab:accuracy}, SIFT+FLANN instead exceeds both SuperPoint pairings on four of the five synthetic datasets, a lead that comes from the translation direction, which short baselines constrain poorly. Which family comes out ahead therefore depends on the metric, and since rover VO and relocalization need both rotation and translation direction, neither score alone predicts how a front end will localize. The datasets with the largest margins also differ from the synthetic ones in more than appearance, so we do not attribute the margins to simulation, although synthetic results alone do not settle the relative performance expected on real rover imagery. In Table~\ref{tab:accuracy}, a SuperPoint pairing is best on every real dataset except S3LI (Vulcano), and the order of the top three on LuPNT (base) and of the top two on LuPNT (high sun) is not resolved.

The ablation attributes most of the testbed lead to the SuperPoint detector and descriptor rather than the matcher. On PolarTraverse and LunaPolaris, SuperPoint with FLANN already exceeds every SIFT combination, although with brute-force matching it falls below SIFT+LightGlue on PolarTraverse. On synthetic data, LightGlue clearly improves SuperPoint over FLANN but barely changes SIFT. Converting SIFT to RootSIFT, the descriptor of LightGlue's SIFT model, changes accuracy little anywhere. Nor does the lead come from more keypoints, as SuperPoint returns fewer than SIFT on every dataset except PolarTraverse, so it lies in the quality of the keypoints and descriptors. This suggests that on other real terrain replacing the classical detector and descriptor matters more than replacing the matcher.

On S3LI (Etna) the SuperPoint lead is widest on pairs far apart (Fig.~\ref{fig:step_auc}). On PolarTraverse, whose pairs are widely spaced, the SuperPoint pairings are more than twice as accurate as any classical combination. On S3LI (Vulcano) the lead of SIFT+LightGlue over the better SuperPoint pairing is borderline across bootstrap seeds, and perturbing the S3LI reference rotations by their modeled error barely changes the order. Evaluating only closely spaced frames would therefore understate the benefit of SuperPoint on real terrain.

\begin{table*}[t]
  \centering
  \caption{\bfseries\boldmath Operating range $\Delta t^*$ [m], the largest median translation with Pose AUC@$5^\circ \geq 50$\,\%. Pose error as in Table~\ref{tab:accuracy}, rotation alone on LunaPolaris and S3LI. ``---'': never. Bold/underline = best/second. FLANN cells: median of five seeds, spread $\leq 0.46$\,m. AUC: area under the curve, FLANN: fast library for approximate nearest neighbors, SP: SuperPoint.}
  \label{tab:operating_range}
  \renewcommand{\arraystretch}{1.1}
  \footnotesize
  \begin{tabular}{|ll|ccc|cc|c|cc|}
\hline
\hdrstrut & & \multicolumn{5}{c|}{\textbf{\boldmath Synthetic}} & \textbf{\boldmath Testbed} & \multicolumn{2}{c|}{\textbf{\boldmath Analogue}} \\
\cline{3-7}\cline{8-8}\cline{9-10}
\hdrstrut & & \multicolumn{3}{c|}{\textbf{\boldmath LuPNT}} & \multicolumn{2}{c|}{\textbf{\boldmath LuSNAR}} & \textbf{\boldmath LunaPolaris} & \multicolumn{2}{c|}{\textbf{\boldmath S3LI}} \\
\cline{3-5}\cline{6-7}\cline{9-10}
\textbf{\boldmath Detector} & \textbf{\boldmath Matcher} & \textbf{\boldmath (base)} & \textbf{\boldmath (cam.)} & \textbf{\boldmath (high sun)} & \textbf{\boldmath (5)} & \textbf{\boldmath (9)} & & \textbf{\boldmath (Etna)} & \textbf{\boldmath (Vulcano)} \\
\hline\hline
SIFT & FLANN & 2.3 & 4.1 & \best{5.3} & \best{3.7} & \best{3.6} & 0.3 & 2.8 & 6.2 \\
SIFT & BruteForce & 1.3 & 1.8 & 1.9 & 1.8 & 2.1 & --- & 0.9 & 3.7 \\
SIFT & LightGlue & 3.0 & 4.6 & \second{5.3} & \second{3.2} & \second{3.0} & 0.3 & 3.8 & \best{6.6} \\
ORB & FLANN & --- & 0.7 & 1.1 & 0.9 & 0.9 & --- & 0.6 & 1.6 \\
ORB & BruteForce & --- & 1.0 & 1.1 & 1.3 & 1.2 & --- & 0.5 & 1.6 \\
AKAZE & FLANN & 1.3 & 1.9 & 2.6 & 2.5 & 2.9 & --- & 1.4 & 2.5 \\
AKAZE & BruteForce & 1.1 & 1.6 & 1.9 & 2.3 & 2.5 & --- & 0.9 & 1.9 \\
BRISK & FLANN & 2.0 & 2.2 & 4.1 & 2.7 & 2.9 & 0.3 & 1.5 & 4.0 \\
BRISK & BruteForce & 1.2 & 1.5 & 1.6 & 1.4 & 1.5 & --- & 0.7 & 2.4 \\
SP & SuperGlue & \second{5.0} & \best{4.8} & 5.2 & 3.0 & 3.0 & \best{0.9} & \best{5.2} & \second{6.3} \\
SP & LightGlue & \best{5.0} & \second{4.8} & 5.2 & 3.0 & 2.9 & \second{0.8} & \second{4.7} & 5.3 \\
\hline
\end{tabular}

\end{table*}

EP and Pose AUC can disagree. EP is highest for SIFT+FLANN on all five synthetic datasets and for SIFT+LightGlue on PolarTraverse, whereas a SuperPoint pairing leads Pose AUC on LuPNT (base) and PolarTraverse. Only on LunaPolaris is EP highest for a SuperPoint pairing. EP counts a match as correct only within about half a pixel of its epipolar line, so it rewards precise keypoint localization. On LuPNT (base), SuperPoint matches lie within a few pixels of their reference correspondence as often as SIFT matches but much less often within one pixel, consistent with the sub-pixel refinement of SIFT keypoints. SuperPoint also yields about twice as many matches, leaving RANSAC more inliers. EP thus measures how precisely matches are localized rather than how well they constrain the pose, so it should not guide the choice of a front end on its own.

SIFT+FLANN is the best classical combination at default settings on every dataset, counting SIFT+LightGlue as learned throughout, though not resolved from BRISK+FLANN on LunaPolaris. It serves as the classical reference, but only with its ratio test or cross-check. With OpenCV's default nearest-neighbor matching it loses about half its accuracy on LuPNT (base), whereas the SuperPoint pairings filter internally and are unaffected. Match filtering is therefore part of a classical front end, since RANSAC alone does not recover the lost accuracy.

Default detector settings understate the classical detectors on low-texture imagery. On subsamples of LuPNT (base) and LunaPolaris, raising the ORB keypoint cap and lowering the AKAZE detection threshold improves every ORB and AKAZE combination. AKAZE, which detects very few keypoints on LunaPolaris at its default threshold, gains the most. With FLANN matching it overtakes SIFT+FLANN on LunaPolaris but stays below both SuperPoint pairings, and on LuPNT (base) it roughly matches SIFT+FLANN. A classical detector should therefore be tuned to the target imagery before it is judged unsuitable for a rover.

Because the best front end differs between datasets, a front end should be chosen from the results on the dataset closest to the target terrain (Table~\ref{tab:accuracy}) and at the rover's frame spacing (Fig.~\ref{fig:step_auc}), not extrapolated from the synthetic results.

\subsection{Accuracy Versus Baseline}
\label{sec:results_gap}

Fig.~\ref{fig:step_auc} and Table~\ref{tab:operating_range} show how pose accuracy depends on the baseline. Accuracy is limited by the translation direction at short baselines, which barely constrain it, and by overlap at long ones, which leave few shared features. Both limits are geometric, so we expect them on other terrain as well. Binned by overlap, the curves mirror Fig.~\ref{fig:step_auc}, since overlap falls as the baseline grows. Where the SuperPoint pairings lead, their advantage lies at low to moderate overlap and generally narrows at high overlap, disappearing on LunaPolaris and reversing on LuPNT (base). On S3LI (Etna) it holds in every bin below 60\,\% overlap.

\subsubsection{Overlap and rotation as difficulty axes}
\label{sec:results_axis_validation}

In the decile test, overlap explains the most variance on average and on three of the five datasets. $\Delta\theta$ comes close but varies most across datasets, and $\bar\alpha$ is weakest, so $\omega$ is our reference axis.

\subsubsection{Per-step accuracy}
\label{sec:results_gap_baseline}

On LuPNT (base), joint Pose AUC@$5^\circ$ peaks at intermediate baselines (Fig.~\ref{fig:step_auc}). At the shortest step, SIFT+FLANN and SP+LightGlue recover the rotation almost perfectly but often miss the translation direction. At the widest step, low overlap makes the rotation fail as well. Scored separately, rotation accuracy is nearly equal up to about 1\,m on LuPNT (base) and LuSNAR (5), so the joint Pose AUC lead of SIFT+FLANN comes from the translation direction. This lead reverses beyond 2\,m on LuPNT (base), where SIFT also loses rotation accuracy, but holds at every tested baseline on LuSNAR (5).

When KLT tracks through every intermediate frame, it is the most accurate front end on all five synthetic datasets under the joint metric (Table~\ref{tab:accuracy}), because each hop between consecutive frames is short. On LunaPolaris it is slightly above the SuperPoint pairings, by a margin we do not test for resolution. On real outdoor terrain it falls below both SuperPoint pairings on S3LI (Etna) and stays close to the other combinations on S3LI (Vulcano) (Table~\ref{tab:rotonly}). Tracked directly between the two frames of a pair, it is less accurate than SIFT+FLANN and both SuperPoint pairings on every dataset, and on LuPNT (base) it collapses at wider steps. Chaining needs every intermediate frame, so it cannot skip frames to save computation, and it cannot verify a revisit, which has no intermediate frames.

On real imagery, the ratio of the best learned to the best classical accuracy grows with the baseline, to about two and a half times on S3LI (Etna) and nearly eight times on LunaPolaris at the widest (Fig.~\ref{fig:step_auc}). On S3LI (Etna) the two are nearly equal up to about 1\,m of translation, while on LunaPolaris it opens from the second step. On low-texture imagery such as LunaPolaris, learned front ends thus help even between closely spaced frames.

\begin{table*}[t]
  \centering
  \caption{\bfseries\boldmath Rotation-only Pose AUC@$5^\circ$ [\%] and stereo VO ATE/L [\%] on LunaPolaris sequences 01 (ceiling lights), 07 (low spotlight) and 10 (low spotlight and headlight). Marks as in Table~\ref{tab:accuracy}. Full pools, not pose-matched. FLANN cells: median of five seeds. \textsuperscript{$\ddagger$}Stopped tracking: path coverage. Sequence 01 repeats the LunaPolaris columns of Tables~\ref{tab:accuracy}, \ref{tab:rotonly} and~\ref{tab:vo}. ATE: absolute trajectory error, ATE/L: ATE as a percentage of path length, AUC: area under the curve, FLANN: fast library for approximate nearest neighbors, SP: SuperPoint, VO: visual odometry.}
  \label{tab:lighting}
  \renewcommand{\arraystretch}{1.1}
  \footnotesize
  \begin{tabular}{|ll|ccc|ccc|}
\hline
\hdrstrut & & \multicolumn{3}{c|}{\textbf{\boldmath Pose AUC@$5^\circ$ [\%]}} & \multicolumn{3}{c|}{\textbf{\boldmath Stereo VO ATE/L [\%]}} \\
\cline{3-5}\cline{6-8}
\textbf{\boldmath Detector} & \textbf{\boldmath Matcher} & \textbf{\boldmath Seq.\,01} & \textbf{\boldmath Seq.\,07} & \textbf{\boldmath Seq.\,10} & \textbf{\boldmath Seq.\,01} & \textbf{\boldmath Seq.\,07} & \textbf{\boldmath Seq.\,10} \\
\hline\hline
SIFT & FLANN & 31.6 & 30.3 & 42.1 & 2\,\%\textsuperscript{$\ddagger$} & 0.6 & 0.8 \\
SIFT & BruteForce & 15.0 & 20.3 & 25.9 & $<$1\,\%\textsuperscript{$\ddagger$} & 11\,\%\textsuperscript{$\ddagger$} & 71\,\%\textsuperscript{$\ddagger$} \\
SIFT & LightGlue & 29.2 & 37.4 & 49.3 & 2\,\%\textsuperscript{$\ddagger$} & \second{0.4} & 1.1 \\
\hline
ORB & FLANN & 9.9 & 8.7 & 14.3 & 13\,\%\textsuperscript{$\ddagger$} & 9\,\%\textsuperscript{$\ddagger$} & 2.1 \\
ORB & BruteForce & 6.7 & 8.6 & 14.5 & 13\,\%\textsuperscript{$\ddagger$} & 9\,\%\textsuperscript{$\ddagger$} & 69\,\%\textsuperscript{$\ddagger$} \\
\hline
AKAZE & FLANN & 12.5 & 21.6 & 28.2 & $<$1\,\%\textsuperscript{$\ddagger$} & 0.6 & \second{0.5} \\
AKAZE & BruteForce & 11.0 & 19.9 & 26.4 & $<$1\,\%\textsuperscript{$\ddagger$} & 43\,\%\textsuperscript{$\ddagger$} & 0.5 \\
\hline
BRISK & FLANN & 30.4 & 26.7 & 36.8 & 38\,\%\textsuperscript{$\ddagger$} & \best{0.4} & \best{0.5} \\
BRISK & BruteForce & 9.0 & 17.7 & 19.8 & 1\,\%\textsuperscript{$\ddagger$} & 56\,\%\textsuperscript{$\ddagger$} & 69\,\%\textsuperscript{$\ddagger$} \\
\hline
SP & SuperGlue & \best{56.9} & \best{41.7} & \best{54.1} & \best{1.3} & 0.6 & 0.6 \\
SP & LightGlue & \second{55.2} & \second{41.4} & \second{53.6} & \second{1.6} & 0.5 & 0.5 \\
\hline
\end{tabular}

\end{table*}

\begin{figure*}[t]
  \centering
  \includegraphics[width=\linewidth]{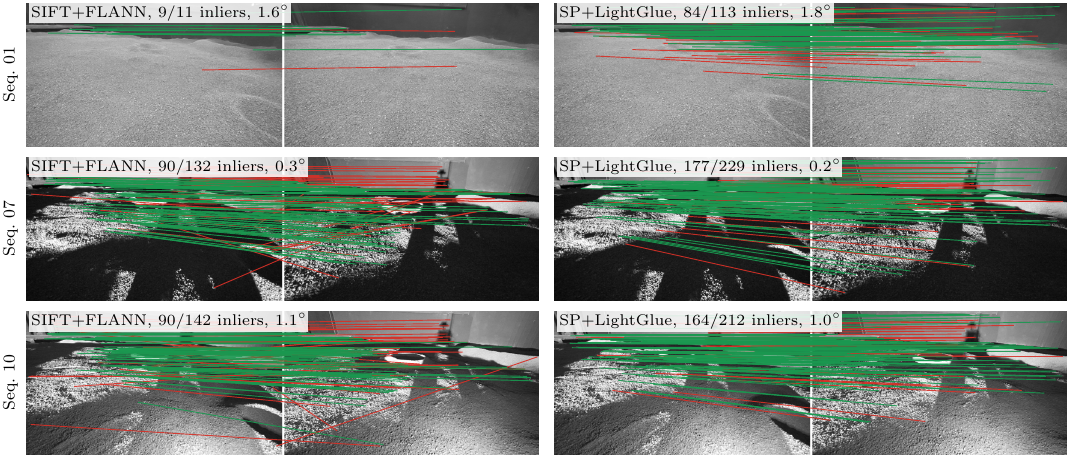}
  \caption{\bfseries\boldmath Matches of SIFT+FLANN (left) and SuperPoint+LightGlue (right), with inlier counts and rotation errors, on a $0.34$\,m pair from sequence 01 (top) and its pose-matched counterparts in 07 (middle) and 10 (bottom). Green = RANSAC inliers, red = outliers. FLANN: fast library for approximate nearest neighbors, RANSAC: random sample consensus, SP: SuperPoint.}
  \label{fig:lighting}
\end{figure*}

\subsubsection{Operating range}
\label{sec:results_range}

The operating range $\Delta t^*$ (Eq.~\eqref{eq:operating_range}) shows the SuperPoint advantage more clearly than the pool averages. On LuPNT (base) and S3LI (Etna), the SuperPoint pairings keep 50\,\% Pose AUC@$5^\circ$ roughly twice as far as any classical combination, and on LunaPolaris about three times as far as any other combination. SIFT-based combinations have the longest range on the LuSNAR scenes, LuPNT (high sun) and S3LI (Vulcano). Because the baseline between matched frames is the driving speed divided by the matching rate, a longer operating range would let a rover drive faster or match fewer frames, although we do not test whether skipping frames preserves VO performance.

\subsection{Camera Effects, Lighting and Cross-Dataset Transfer}
\label{sec:results_pairwise}

Four comparisons change the appearance of the imagery: LuPNT with camera effects or a higher sun, LuPNT (base) against S3LI (Etna), and three lightings along one LunaPolaris trajectory (Table~\ref{tab:lighting} and Fig.~\ref{fig:lighting}). LuPNT (base) against S3LI (Etna) also changes the scene and pair geometry, so it does not isolate appearance. Lighting affects front ends with classical detectors more than the SuperPoint pairings, so a front end should be evaluated under the illumination expected during operations.

Camera effects lower EP but not classical pose accuracy. On pairs with the same frame indices in both collections, LuPNT (cam.) has lower mean EP than LuPNT (base) for every combination except the two ORB pairings. It has higher joint Pose AUC@$5^\circ$ for all but BRISK+FLANN, which is unchanged, and the SuperPoint pairings, which lose slightly.
The camera-effects run retraces the base path within a few centimeters, so the differences come from the camera effects rather than pair geometry, although it is unclear why Pose AUC rises while EP falls. Modeled camera effects of this kind therefore do not by themselves reproduce the accuracy loss seen on the real datasets.

Every combination is less accurate on S3LI (Etna) than on LuPNT (base) in rotation-only Pose AUC@$5^\circ$. The FLANN combinations already trail at the shortest step, so the deficit is not only a wide-baseline effect. The two pools also differ in pair geometry, scene, resolution and ground truth, so we do not attribute it to appearance alone. Absolute accuracy measured in simulation is therefore likely optimistic for a rover on real terrain, whatever the cause.

LunaPolaris sequences 01 (ceiling lights), 07 (a single low spotlight) and 10 (the spotlight and a rover headlight) follow one trajectory. The spotlight raises the RMS image contrast by almost half and saturates a small fraction of the pixels. Its shadows cover about an eighth of the image on 07 but almost none on 10, which adds the headlight.

The full pools (Table~\ref{tab:lighting}) hide the spotlight's gain for the combinations with classical detectors by mixing lighting with geometry. A frame step spans a longer baseline on 07, where several combinations, including the SuperPoint pairings, are less accurate. On pairs matched in pose across the three sequences, every combination with a classical detector gains significantly under both spotlight lightings, also at a looser pose tolerance, while the SuperPoint pairings show no resolved change, so the gap to SuperPoint narrows. The significant SuperPoint fall on the full 07 pool is consistent with the longer baselines there.

Under the spotlight every detector returns more keypoints and SIFT+FLANN keeps more matches, although few keypoints fall in the shadows. On the screened pose-matched pairs (one shown in Fig.~\ref{fig:lighting}), SIFT+FLANN recovers the rotation within $5^\circ$ on just over half the pairs under the ceiling lights but on nearly all under the spotlight, while SP+LightGlue does so on nearly every pair under each lighting. This is consistent with low-elevation light shading the relief into texture the classical detectors can use, whereas diffuse light gives flat images and deep shadow shows almost none. On LuPNT, a $45^\circ$ sun also raises the accuracy of the combinations with classical detectors while the SuperPoint pairings barely change, although camera effects at $15^\circ$ already give most of this gain for all but BRISK (Table~\ref{tab:accuracy}). On the Moon, where sunlight is directional, this suggests classical matching is weakest in deep shadow and under a near-overhead sun that flattens the shading, neither of which we isolate.

\begin{figure*}[t]
  \centering
  \includegraphics[width=\linewidth]{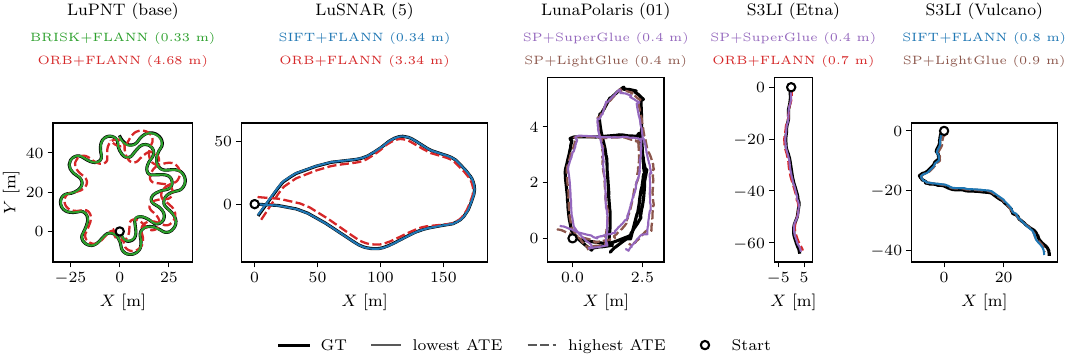}
  \caption{\bfseries\boldmath Top-down stereo VO trajectories, SE(3)-aligned to GT, of the lowest (solid) and highest (dashed) ATE combination that tracks the whole path, on five datasets covering the three tiers. The S3LI sites show their scored first 10\,\%. ATE: absolute trajectory error, FLANN: fast library for approximate nearest neighbors, GT: ground truth, SP: SuperPoint, VO: visual odometry.}
  \label{fig:vo_traj}

  \vspace{4mm}
  \centering
  \captionof{table}{\bfseries\boldmath VO ATE [m] with stereo depth on every dataset and with GT depth on the synthetic datasets. FLANN cells: median of five seeds (median spread 17\,\%). \textsuperscript{$\ddagger$}Stopped tracking: path coverage. \emph{Tracked}: datasets with $\geq 90$\,\% path coverage in stereo VO. Bold/underline = best/second per column and block. ATE: absolute trajectory error, FLANN: fast library for approximate nearest neighbors, GT: ground truth, SP: SuperPoint, VO: visual odometry.}
  \label{tab:vo}
  \renewcommand{\arraystretch}{1.1}
  \footnotesize
  
  \begin{tabular}{|ll|ccc|cc|c|cc|c|}
  \hline
  \hdrstrut & & \multicolumn{5}{c|}{\textbf{\boldmath Synthetic}} & \textbf{\boldmath Testbed} & \multicolumn{2}{c|}{\textbf{\boldmath Analogue}} &  \\
  \cline{3-7}\cline{8-8}\cline{9-10}
  \hdrstrut & & \multicolumn{3}{c|}{\textbf{\boldmath LuPNT}} & \multicolumn{2}{c|}{\textbf{\boldmath LuSNAR}} & \textbf{\boldmath LunaPolaris} & \multicolumn{2}{c|}{\textbf{\boldmath S3LI}} &  \\
  \cline{3-5}\cline{6-7}\cline{9-10}
  \textbf{\boldmath Detector} & \textbf{\boldmath Matcher} & \textbf{\boldmath (base)} & \textbf{\boldmath (cam.)} & \textbf{\boldmath (high sun)} & \textbf{\boldmath (5)} & \textbf{\boldmath (9)} & & \textbf{\boldmath (Etna)} & \textbf{\boldmath (Vulcano)} & \textbf{\boldmath Tracked} \\
  \hline\hline
  \multicolumn{11}{|l|}{\textit{Stereo VO}} \\
  \hline
  SIFT & FLANN & \second{0.56} & \second{1.21} & \second{0.05} & \best{0.34} & \second{0.30} & \textsuperscript{$\ddagger$}\,2\,\% & 0.5 & \best{0.8} & 7/8 \\
  SIFT & LightGlue & 0.66 & \best{1.20} & \best{0.04} & 0.42 & \best{0.28} & \textsuperscript{$\ddagger$}\,2\,\% & 0.4 & \second{0.8} & 7/8 \\
  ORB & FLANN & 4.68 & 2.94 & 1.28 & 3.34 & 1.92 & \textsuperscript{$\ddagger$}\,13\,\% & 0.7 & \textsuperscript{$\ddagger$}\,1\,\% & 6/8 \\
  AKAZE & FLANN & 0.94 & 1.31 & 0.05 & \second{0.35} & 0.47 & \textsuperscript{$\ddagger$}\,$<$1\,\% & 0.5 & \textsuperscript{$\ddagger$}\,1\,\% & 6/8 \\
  BRISK & FLANN & \best{0.33} & 1.45 & 0.15 & 0.55 & 1.32 & \textsuperscript{$\ddagger$}\,38\,\% & 0.6 & 0.9 & 7/8 \\
  SP & SuperGlue & 0.68 & 1.66 & 0.65 & 0.47 & 0.92 & \best{0.4} & \best{0.4} & 0.9 & 8/8 \\
  SP & LightGlue & 0.93 & 1.54 & 0.61 & 1.24 & 0.77 & \second{0.4} & \second{0.4} & 0.9 & 8/8 \\
  \hline
  \multicolumn{7}{|l|}{\textit{GT-depth VO}} & \multicolumn{4}{c}{} \\
  \cline{1-7}
  SIFT & FLANN & 0.50 & 1.00 & \best{0.14} & \best{0.31} & 1.09 & \multicolumn{4}{c}{} \\
  SIFT & LightGlue & 0.55 & 1.00 & \second{0.15} & 0.37 & 1.07 & \multicolumn{4}{c}{} \\
  ORB & FLANN & 2.47 & 1.36 & 0.92 & 2.19 & 2.62 & \multicolumn{4}{c}{} \\
  AKAZE & FLANN & 1.14 & 1.07 & 0.29 & \second{0.32} & \second{0.86} & \multicolumn{4}{c}{} \\
  BRISK & FLANN & 0.53 & 1.05 & 0.17 & 0.58 & 1.40 & \multicolumn{4}{c}{} \\
  SP & SuperGlue & \second{0.39} & \best{0.88} & 0.31 & 0.87 & \best{0.47} & \multicolumn{4}{c}{} \\
  SP & LightGlue & \best{0.28} & \second{0.91} & 0.26 & 0.61 & 1.10 & \multicolumn{4}{c}{} \\
  \cline{1-7}
  \end{tabular}
\end{figure*}

\begin{figure*}[t]
  \centering
  \includegraphics[width=\linewidth]{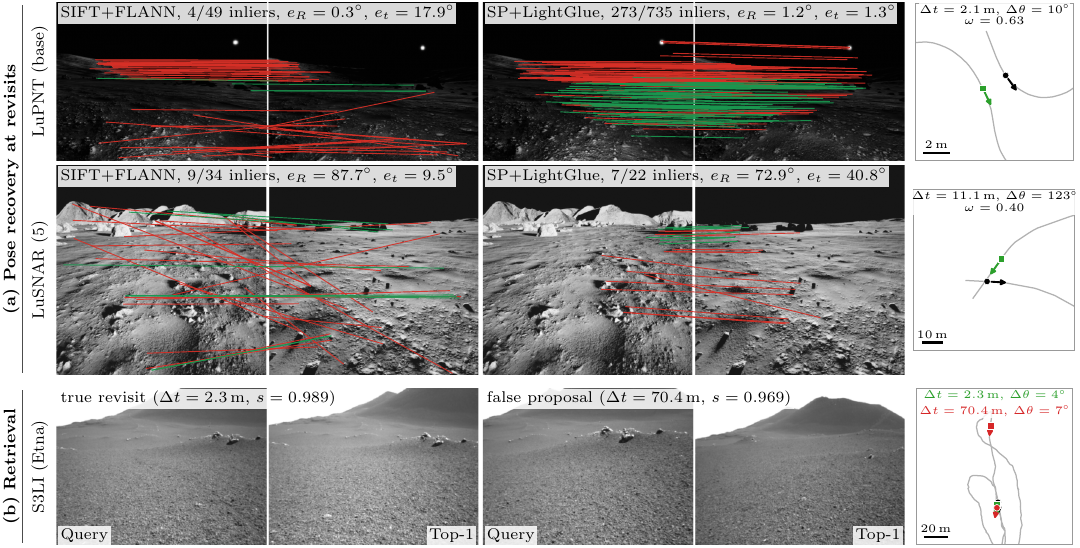}
  \caption{\bfseries\boldmath Loop-closure examples. (a) Matches of SIFT+FLANN (left) and SuperPoint+LightGlue (right), with RANSAC inliers (green), outliers (red) and rotation and translation-direction errors $e_R$, $e_t$. The pairs are a same-direction revisit on LuPNT (base) and, on LuSNAR (5), a pair seen from directions $123^\circ$ apart that shares $40\,\%$ of the view. (b) S3LI (Etna) DINOv2 retrieval: one query whose top-1 is a true revisit (left) and one whose top-1 is a false proposal (right). Top-down maps at far right. FLANN: fast library for approximate nearest neighbors, RANSAC: random sample consensus, SP: SuperPoint.}
  \label{fig:loop_qual}

  \vspace{4mm}
\centering
  \captionof{table}{\bfseries\boldmath Revisit structure (at $R=1$\,m, $2$\,m on LuPNT (base)) and DINOv2 retrieval within radius $R$. Rows with fewer than three components are exploratory. S3LI frames with GNSS jumps are excluded. ``---'' = no recall at 100\,\% precision. AUPRC: area under the precision-recall curve, Comp.: revisit components (revisited places), R@1: recall of the top match, R@100P: recall before the first false positive.}
  \label{tab:loop_closure}
  \renewcommand{\arraystretch}{1.1}
  \footnotesize
  \setlength{\tabcolsep}{2pt}
  \begin{tabular}{|l|ccccl|cccccccc|}
  \hline
  \hdrstrut & \multicolumn{5}{c|}{\textbf{\boldmath Revisits}} & \multicolumn{8}{c|}{\textbf{\boldmath Retrieval}} \\
  \cline{2-6}\cline{7-14}
  \hdrstrut \textbf{\boldmath Dataset} & \textbf{\boldmath Pairs} & \textbf{\boldmath Comp.} & \tab{c}{c}{\textbf{\boldmath Median}\\\textbf{\boldmath $\Delta\theta$\,[$^\circ$]}} & \tab{c}{c}{\textbf{\boldmath Fraction}\\\textbf{\boldmath $<30^\circ$}} & \textbf{\boldmath Type} & \textbf{\boldmath $R$} & \textbf{\boldmath Queries} & \textbf{\boldmath Comp.} & \tab{c}{c}{\textbf{\boldmath R@1}\\\textbf{\boldmath (causal)}} & \tab{c}{c}{\textbf{\boldmath R@1}\\\textbf{\boldmath (offline)}} & \tab{c}{c}{\textbf{\boldmath AUPRC}\\\textbf{\boldmath (causal)}} & \tab{c}{c}{\textbf{\boldmath AUPRC}\\\textbf{\boldmath (offline)}} & \tab{c}{c}{\textbf{\boldmath R@100P}\\\textbf{\boldmath (causal)}} \\
  \hline\hline
    \multirow{2}{*}{LuPNT (base)} & \multirow{2}{*}{596} & \multirow{2}{*}{3} & \multirow{2}{*}{5.7} & \multirow{2}{*}{1.00} & \multirow{2}{*}{\tab{l}{l}{low-angle}} & 2\,m & 32 & 3 & 0.250 & 0.212 & 0.023 & 0.017 & --- \\
     &  &  &  &  &  & 5\,m & 450 & 1 & 0.324 & 0.306 & 0.082 & 0.080 & --- \\
    \hline
    \multirow{3}{*}{S3LI (Etna)} & \multirow{3}{*}{220{,}069} & \multirow{3}{*}{5} & \multirow{3}{*}{41.3} & \multirow{3}{*}{0.41} & \multirow{3}{*}{\tab{l}{l}{wide-angle}} & 1\,m & 246 & 5 & 0.142 & 0.124 & 0.012 & 0.010 & --- \\
     &  &  &  &  &  & 2\,m & 423 & 4 & 0.232 & 0.232 & 0.064 & 0.066 & --- \\
     &  &  &  &  &  & 5\,m & 630 & 3 & 0.357 & 0.359 & 0.195 & 0.193 & 0.014 \\
    \hline
    \multirow{3}{*}{S3LI (Vulcano)} & \multirow{3}{*}{83{,}027} & \multirow{3}{*}{2} & \multirow{3}{*}{62.0} & \multirow{3}{*}{0.36} & \multirow{3}{*}{\tab{l}{l}{wide-angle}} & 1\,m & 632 & 2 & 0.123 & 0.142 & 0.037 & 0.045 & 0.002 \\
     &  &  &  &  &  & 2\,m & 968 & 2 & 0.230 & 0.302 & 0.123 & 0.187 & 0.014 \\
     &  &  &  &  &  & 5\,m & 1348 & 2 & 0.322 & 0.436 & 0.196 & 0.308 & 0.061 \\
  \hline
  \end{tabular}
\end{figure*}

\subsection{Visual Odometry}
\label{sec:results_vo}

Pairwise accuracy differences carry over to VO (Table~\ref{tab:vo}) only where the front end is stressed. Elsewhere, with metric depth from ground truth or stereo, nearly every front end tracks accurately. With ground-truth depth, about half the ATE differences between front ends are below the variation between repeated runs.
Three differences between VO and the pairwise evaluation are consistent with this. First, VO matches only consecutive frames, where both families recover rotation almost equally. Second, metric depth makes PnP well posed at zero baseline, removing the translation-direction ambiguity that dominates pairwise error at short steps. Third, VO loses a frame only when too few matches or inliers with depth remain, so match support between consecutive frames matters more than wide-baseline accuracy.

\subsubsection{Visual odometry with ground-truth depth}

PnP on ground-truth depth gives metric VO on the synthetic datasets, and all eleven combinations stay at or below about $0.7\,\%$ ATE/L.

With ground-truth depth, only ORB+FLANN stands out among the tabulated combinations. It has the highest ATE in every ground-truth-depth column, by more than its seed spread everywhere except LuPNT (high sun) and LuSNAR (5). The larger LuSNAR leads that involve a SuperPoint run are within the variation between GPU reruns and not resolved. The weakness of ORB+FLANN is consistent with the default ORB keypoint cap, since raising it improves every ORB combination in the pairwise test. Beyond this, ground-truth-depth VO barely separates the front ends.

Front ends that stay accurate over longer baselines also tend to have lower ATE/L. With ground-truth depth, the Spearman correlation between operating range and ATE/L over the synthetic combinations and datasets is $\rho=-0.56$. With stereo depth it weakens to $\rho=-0.41$, because the VO error of the learned front ends no longer follows their pairwise accuracy. A short operating range thus flags a front end likely to drift, but the pairwise benchmark is a screen, not a substitute for running VO. The correlation is in-sample and sensitive to the 50\,\% threshold, so it indicates a trend rather than a calibrated prediction.

\subsubsection{Stereo visual odometry}
\label{sec:results_stereo_vo}

On LunaPolaris, the lidar reference depth only describes pair geometry, not stereo VO.

On the S3LI sites stereo VO is mostly a tie, whereas on LunaPolaris only the SuperPoint pairings stay on track. On Etna every tabulated combination tracks every frame, all but ORB+FLANN tie, and every trajectory stays within 0.8\,m RMS of the ground truth (Fig.~\ref{fig:vo_traj}). On Vulcano, ORB+FLANN and AKAZE+FLANN stop within the first 1\,\% of the path, where the camera pitches toward the sea and sky, and the others tie. On LunaPolaris sequence 01 every other combination stops within 38\,\% of the path. Under the spotlight (Table~\ref{tab:lighting}), SIFT+FLANN, SIFT+LightGlue, BRISK+FLANN and AKAZE+FLANN complete both paths, ORB+FLANN and AKAZE+BruteForce only that of 10, and the other brute-force combinations stop on both. This suggests that on low-texture terrain VO with classical detectors is limited by how much texture the lighting reveals, whereas the SuperPoint pairings track under all three lightings.

Across the eight datasets, only the SuperPoint pairings keep tracking on all of them in stereo VO (Table~\ref{tab:vo}, \emph{Tracked}). Front-end selection therefore depends on the task. Frame-to-frame VO needs enough reliable matches to keep tracking, while wider-baseline localization also needs an accurate relative pose, so pairwise accuracy should be read alongside tracking coverage.

With stereo depth, the SuperPoint pairings lose the unresolved lead they hold with ground-truth depth on three of the synthetic datasets (Table~\ref{tab:vo}), and show no resolved VO accuracy advantage over the SIFT pairings. Every combination tracks the whole path to within a few meters (Fig.~\ref{fig:vo_traj}). On LuPNT the stereo ATE of the SuperPoint pairings is 1.7 to 3.3 times that with ground-truth depth, consistent with stereo depth error, rather than matching, limiting their accuracy. The VO advantage of SuperPoint therefore lies in staying on track, as on LunaPolaris, rather than in trajectory accuracy.

\subsection{Loop Closure}
\label{sec:results_slam}

No evaluated front end handles every kind of revisit. The two tested combinations rarely recover the pose on revisits seen from widely different directions, even with partial overlap.

\subsubsection{Revisit structure}

Revisits are counted at $R=1$\,m ($R=2$\,m for LuPNT (base), which otherwise has no revisits). Four datasets have fewer than three revisit components. LuSNAR (5) has a single crossing, LuSNAR (9) has no revisit within 10\,m, S3LI (Vulcano) has two, and the revisits of LunaPolaris, where the rover laps an arena of about 3 by 6\,m (Fig.~\ref{fig:vo_traj}), overlap along the whole path and join into a single component of places that cannot be separated. Retrieval is therefore evaluated on LuPNT (base) and S3LI (Etna), with S3LI (Vulcano) exploratory.

Revisit rotation varies widely between datasets (Table~\ref{tab:loop_closure}). On both S3LI sites most revisit pairs exceed $30^\circ$, so a viewing-angle limit as in MSLS~\cite{warburg_msls_2020} would discard many of them.

\subsubsection{Pose recovery at revisits}

On the same-direction retraces of LuPNT (base), both SuperPoint pairings reach about five times the Pose AUC@$5^\circ$ of SIFT+FLANN, the best classical combination (Fig.~\ref{fig:loop_qual}, top row). These three recover the rotation alone far more often, so SIFT+FLANN, whose candidate matches RANSAC mostly rejects, loses mainly the translation direction. The other classical combinations also lose the rotation. For a rover retracing its path, this suggests a SuperPoint front end would supply many more accurate loop-closure constraints than a classical one.

Being close in position does not make two frames a usable revisit. The revisit pairs of LuSNAR (5) all come from one crossing where the headings differ by $123^\circ$, so the two views share no overlap to match. Near that crossing, frames several meters apart share part of the view but from directions too different to match, and both SIFT+FLANN and SP+LightGlue almost never recover the pose (Fig.~\ref{fig:loop_qual}, second row). A radius-only definition still counts both kinds of pair as revisits, so loop-closure evaluation should define revisits by overlap and viewing direction, not position alone.

\begin{figure}[!t]
  \centering
  \includegraphics[width=\linewidth]{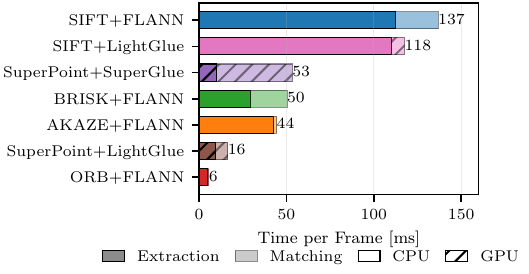}
  \caption{\bfseries\boldmath Incremental cost per frame [ms] on LuPNT (base). Hatched bars run on the GPU. CPU: central processing unit, FLANN: fast library for approximate nearest neighbors, GPU: graphics processing unit.}
  \label{fig:timing}
\end{figure}

\subsubsection{Retrieval}

R@100P is below 0.07 in every row (Table~\ref{tab:loop_closure}), so proposals from this descriptor need verification before they enter a pose graph. Fig.~\ref{fig:loop_qual}(b) shows why: a false proposal 70\,m from its query, on visually similar terrain, scores almost as high as a true revisit. Lunar terrain is likely at least as self-similar as these volcanic slopes, so we expect the same near-duplicate proposals there. A DINOv3~\cite{simeoni_dinov3_2025} CLS-token descriptor has lower AUPRC in all eight rows, so we report DINOv2. AUPRC measures how well true revisits rank above false proposals, which corrupt a pose graph. The CLS token, rather than the terrain, may be the limit, since AnyLoc's patch-token VLAD aggregation on the same backbone improves recall@1 over it in unstructured environments~\cite{keetha_anyloc_2023}. These results do not show that appearance-based retrieval fails on lunar terrain.

\subsection{Computational Cost}
\label{sec:results_timing}

SP+LightGlue is the fastest learned combination (Fig.~\ref{fig:timing}) and, at 16\,ms, faster than every timed classical combination except ORB+FLANN, because CPU extraction dominates the classical cost. It is about three times as fast as the SuperGlue pairing, as LightGlue exits early and prunes points~\cite{lindenberger_lightglue_2023}. On this hardware SP+LightGlue costs no more per frame than most classical ones, although rover-class processors, which we do not test, could differ. Retrieval adds one DINOv2 descriptor per keyframe, about the cost of one SP+LightGlue frame, plus a negligible similarity search.

\section{Conclusion}
\label{sec:conclusion}

Lunar rovers will localize with cameras, yet feature matchers are developed on terrestrial scenes and planetary work compares them on orbital registration or within complete SLAM systems, not between rover frames. We therefore benchmark eleven detector--matcher combinations on nine synthetic, lunar-testbed and terrestrial-analogue datasets, and a KLT tracker, scoring the pose accuracy of image pairs against their baseline and the trajectory error of the resulting VO.

Scored by rotation alone, the SuperPoint pairings outperform SIFT+FLANN on both lunar testbeds and one analogue site and stay accurate out to longer baselines, while the differences on the synthetic datasets are smaller or unresolved. Scored on rotation and translation direction jointly, SIFT+FLANN exceeds both SuperPoint pairings on four synthetic datasets, because its translation direction is more accurate at short baselines. Synthetic results alone therefore do not establish relative performance on real rover imagery. Modeled camera effects lower EP but not classical Pose AUC, so they do not reproduce the loss seen on real imagery. The ablation attributes most of the testbed lead to the SuperPoint detector and descriptor rather than the learned matcher. A KLT tracker chained through every frame is the most accurate front end on the synthetic datasets under the joint metric and slightly above the SuperPoint pairings on LunaPolaris, but not at the Etna analogue site, and it cannot skip frames or verify revisits.

In VO, the front ends differ more in whether they stay on track than in accuracy. With ground-truth depth, about half their ATE differences are within run-to-run variation, and with stereo depth no consistent SuperPoint advantage is resolved, but only the SuperPoint pairings keep tracking on all eight sequential datasets, including under the diffuse ceiling lighting of LunaPolaris. Pairings with classical detectors depend on lighting while SuperPoint changes little: a low spotlight makes them more accurate than diffuse light on the same testbed views, consistent with more contrast and detected keypoints. SuperPoint thus likely matters most on uniformly lit regolith, such as under a near-overhead sun, and possibly in deep polar shadow. At revisits, SuperPoint reaches about five times the Pose AUC of SIFT+FLANN on same-direction retraces, but SIFT+FLANN and SP+LightGlue almost never recover the pose between views from widely different directions, even with partial overlap, and DINOv2 retrieval proposals need geometric verification.

For a rover with a GPU, we recommend SuperPoint with LightGlue, the faster SuperPoint pairing, about as accurate as SuperGlue. Without a GPU, SIFT+FLANN has the highest classical Pose AUC at default settings, although AKAZE with a lower detection threshold surpasses it on a LunaPolaris subsample, and where it keeps tracking, its VO is not consistently less accurate than that of SuperPoint.

Future work will tune the classical detectors per dataset, sweep the sun elevation in simulation, select pairs by measured geometry, adapt the learned matchers to analogue imagery, evaluate complete SLAM on revisits defined by overlap, and build a camera-independent reference for S3LI. We will also extend the benchmark to real lunar surface imagery from Chang'e-3 and Apollo.

\acknowledgements

We gratefully acknowledge Blue Origin for funding this project and for valuable discussions that contributed to its development. This manuscript benefited from the use of AI-based assistants, Claude Opus and Claude Sonnet, which were employed for coding assistance and revising. All final content was reviewed and edited by the authors.

\bibliographystyle{IEEEtran}
\bibliography{refs}

\thebiography

\begin{biographywithpic}
	{Guillem Casadesus Vila}{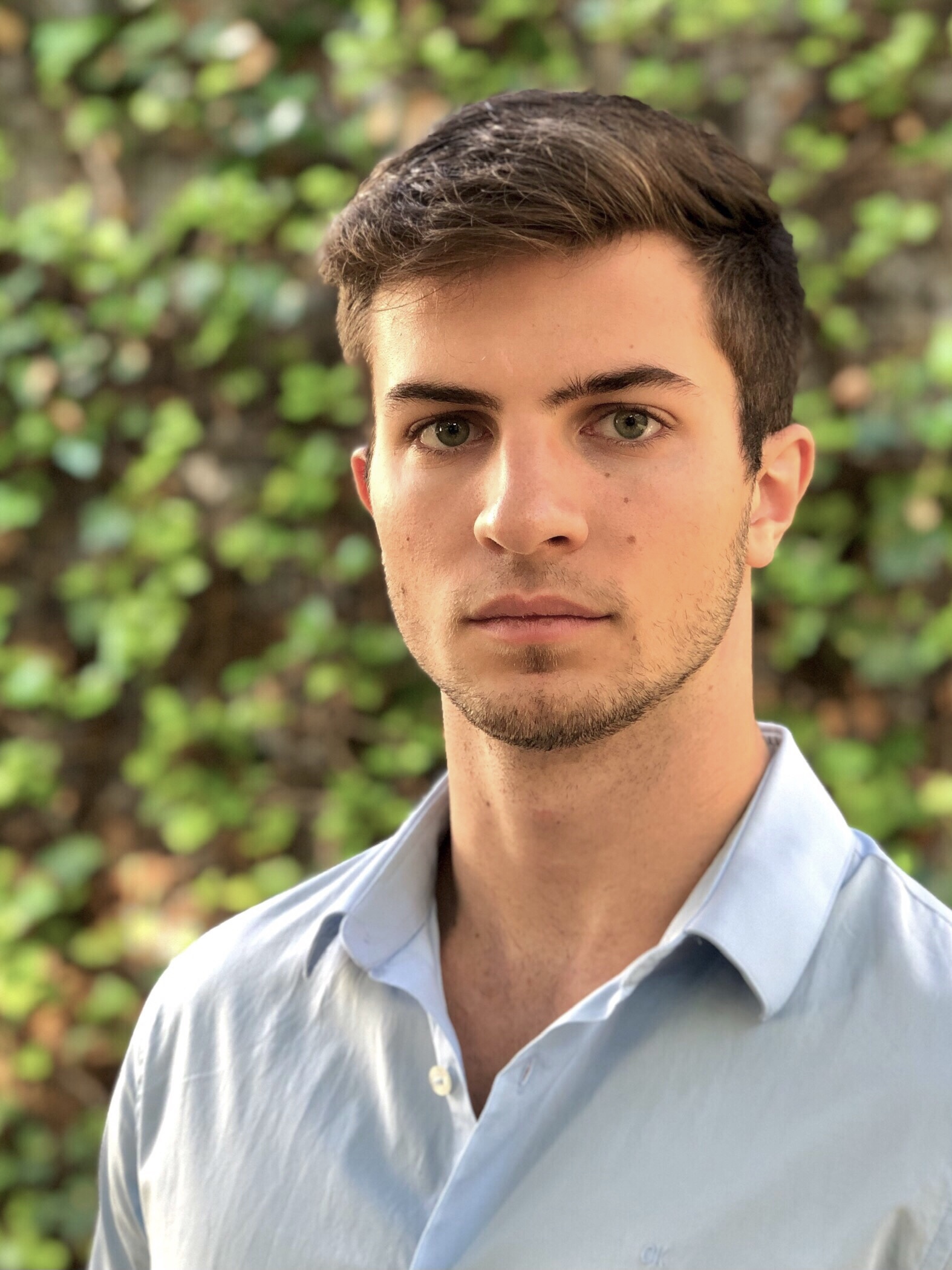} is a Ph.D. candidate in the Department of Aeronautics and Astronautics at Stanford University. He received his B.Sc. degree in Aerospace and Telecommunications Engineering in 2022 from the Universitat Polit\`ecnica de Catalunya (UPC) under the CFIS program. His research interests include robotic space exploration and space navigation and communication.
\end{biographywithpic}

\begin{biographywithpic}
	{Grace Gao}{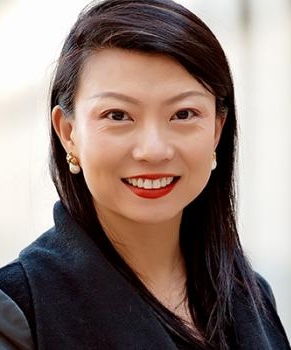} is an associate professor in the Department of Aeronautics and Astronautics at Stanford University. Before joining Stanford University, she was an assistant professor at University of Illinois at Urbana-Champaign. She obtained her Ph.D. degree at Stanford University. Her research is on robust and secure positioning, navigation, and timing with applications to manned and unmanned aerial vehicles, autonomous driving cars, as well as space robotics.
\end{biographywithpic}
\vfill

\end{document}